\documentclass[runningheads]{llncs}
\usepackage{bibnames}
\usepackage{graphicx}
\usepackage{placeins}
\begin{document}
\title{Exploring Knowledge Distillation of a Deep Neural Network for Multi-Script identification } 
\titlerunning{Distilling Deep Neural Network}
%
\author{Shuvayan Ghosh Dastidar\inst{1} \and
Kalpita Dutta\inst{1} \and
Nibaran Das\inst{1} \and
Mahantapas Kundu\inst{1} \and
Mita Nasipuri\inst{1}}
\authorrunning{Jadavpur University}
%
\institute{Jadavpur University,Kolkata, India \and
\email{kalpitadutta.cse.rs@jadavpuruniversity.in} }
\maketitle              
\begin{abstract}
Multi-lingual script identification is a difficult task consisting of different language with complex backgrounds in scene text images. According to the current research scenario,  deep neural networks are employed as teacher models to train a smaller student network by utilizing the teacher model's predictions. This process is known as dark knowledge transfer. It has been quite successful in many domains where the final result obtained is unachievable through directly training the student network with a simple architecture. In this paper, we explore dark knowledge transfer approach using long short-term memory(LSTM) and CNN based assistant model and various deep neural networks as the teacher model, with a simple CNN based student network, in this domain of multi-script identification from natural scene text images. We explore the performance of different teacher models and their ability to transfer knowledge to a student network. Although the small student network's limited size, our approach obtains satisfactory results on a well-known script identification dataset CVSI-2015.

\end{abstract}

\section{Introduction}
Multi-lingual script identification is a crucial task in case of scene text recognition. Scene text in the wild consists of multiple languages and therefore, accurate script identification has become an indispensable part of the scene text OCR systems. Text recognition from natural scene images is relatively complicated, which consists of different textures, contrasts, random brightness and saturation. Owing to the difficulties in this domain, it has sought the attention of many researchers with pattern recognition and computer vision expertise.\\
Deep Neural Networks usually have a vast number of parameters and memory overhead which makes them unsuitable for mobile uses. Distilling the knowledge from the deep neural network and transferring the latter in a smaller student network with much fewer parameters can make it suitable for mobile uses without much loss in the information of the deep or teacher neural network.
\subsection{Motivation} We have applied a new methodology to classify multilingual scripts from natural scene images. 

In this paper, we have used the concept of knowledge transfer~\cite{hinton2015distilling} from deep neural networks to a smaller network, where the learnt features are transferred to the small model with much less parameters.
The main advantage of this concept is less memory overhead for the task of multi-lingual script identification leading to much less inference times which facilitates the incorporation of the models in small mobile devices where computation power is limited. Along with this, we also experiment with various neural networks of different architectures as the teacher network and study the effect of knowledge transfer of learnt features in the student network. With increasing distance between the parameters of the teacher network and the student network, proper distillation of knowledge cannot be obtained. So an auxiliary intermediate assistant model is employed to increase the efficiency of knowledge transfer.
Through series of experiments and ablation studies, we prove that introduction of the assistant model improves the knowledge transfer from the deep teacher network. The parameters of the intermediate assistant network are less than the teacher model but more than the student network. The key contributions of the paper are:
\begin{itemize}
    \item Distilling  a state of the art deep neural network in the field of  multi-script identification from natural scene text images using three steps of training comprising of the teacher, assistant and the student model. We believe this concept of knowledge distillation for multi-lingual script identification from natural scene images has not been introduced before and this is the first work.
    \item Using an assistant model to distill the knowledge learnt by the teacher network and the transfer the same in a smaller student network. The assistant network consists of conv-nets and a bidirectional stacked LSTM~\cite{breuel2017high} for knowledge transfer between the teacher and the student network, both being convolutional neural networks. It has been inferred that the assistant model helps in boosting the student model's accuracy by bridging the gap between the teacher and the student network and thus diminishing the distance between the parameters of the same.
    \item Exploring different deep neural networks employed as the teacher model and comparing their performance based on the transfer of knowledge in the student network. Inception-Net V3, Efficient-Net B4 and Vgg19 have been used as a teacher model with one assistant convolutional-LSTM based model and the final inference is done on a 4 layer convolutional student network. All of the teacher networks mentioned above have varying number of parameters, much more than the assistant and the student models. 
\end{itemize}

The remaining paper is written as - Section 2 is concerned with some previous works and Section 3 briefly presents the proposed method. Section 4 describes some experimental results, and Section 5 ends the paper with the conclusion and future works.

\section{Related Study}

Earlier papers used ensemble methods for model compression~\cite{huang2017densely,jin2019knowledge}. Distillation of knowledge from a teacher network and transferring it to a student network to mimic the teacher network is a basic fundamental concept of knowledge distillation. The first proposed concept of knowledge distillation~\cite{hinton2015distilling} introduces the concept of compressing the knowledge of a more in-depth or
larger model to a single computational efficient neural network. It has introduced the concept of dark knowledge transfer from a deep teacher network to a smaller student network by taking the softmax of the results of the teacher network with a specific temperature value and calculating loss between it and the predicted outputs of the student network. They validated their findings by running on MNIST dataset and, JFT dataset by google and other speech recognition tasks. Since then, knowledge distillation has progressed a lot, and adversarial methods~\cite{NEURIPS2018_019d385e,xu2018training} also have utilized for modelling knowledge transfer between teacher and student. After this study, extensive research has conducted on knowledge distillation. In the paper~\cite{romero2014fitnets} has introduced the transfer of a hidden activation output and other has proposed transferring attention information as knowledge~\cite{zagoruyko2016paying}. 

Article~\cite{yim2017gift} has briefly described the advantages and efficiency of the knowledge distillation. It has described importance of knowledge transfer from teacher to student model using distilled knowledge. They have compared two student deep neural networks trained with teacher network and without teacher model with same size.They have proposed a method of transferring the distilled knowledge between two layers to shows three important points. The student model is more efficient than the original model and it also outperform the original model which is trained from scratch. The student network understand the flow of solving the problem and it start learning with good initial weights. It can learnt and optimized faster than original or normal deep neural network.This paper proves that, the student model reports better efficiency than a normal network without a teacher model.They have compared various knowledge transfer techniques with a normal network without any teacher model for knowledge transfer.They have learned their model with two main condition. First, the teacher model must pretrained with some different dataset and second condition is the teacher model is shallower or deeper than the student model. Their approach contains two step training. 

Article~\cite{chen2017learning} has portrayed an architecture to learn multi class object detection models using knowledge distillation. They have proposed a weighted cross entropy loss for classification task to accounts for the imbalanced in the misclassification for the background class as opposed to object classes. A teacher has bounded regression loss for knowledge transfer and it also adapt layers for hint learning. It allows a student to learn better from the distribution of neurons in the intermediate teacher layers. Using multiple large-scale public benchmarks they have made an empirical evaluation. They have presented the behavior of their framework by relating it to the generalization and under-fitting problems.  They have adopted Faster-RCNN to detect object. They have learnt hint based learning to the similar feature representation of a student model with a teacher model. General knowledge transfer models have a large pre-trained teacher model to train a smaller student model. But article~\cite{mirzadeh2020improved} has proper knowledge transfer is not obtained with increase in gap of the parameters of student and teacher model. A teacher can successfully transfer its knowledge to the students up to a certain size, not too small, they have proposed a multi-step knowledge distillation method, which employs an intermediate-sized teacher assistant network to minimize the gap of the two models - teacher and student. They have studied the effect of teacher assistant model size and has extended the structure to multi-step distillation. First, the TA network has distilled the knowledge from the teacher model. Then, the TA acts as a teacher model to trains the student model via knowledge distillation. Also some other paper ~\cite{gomez2017improving} has focused on assistant network.      
There are some conflicting views regrading whether a good teacher always teaches a student well. Some times if the student model capacity is too low, then knowledge distillation cannot succeed to mimic the teacher model~\cite{cho2019efficacy}.They have presented an approach to mitigate this issue by early stopping teacher training to improve a solution more amenable for the student model. 

In ~\cite{son2020densely} have used densely guided knowledge distillation technique using multiple teacher assistant network. which progressively decrease the model size to bridge the gap efficiently between teacher and student networks.To stimulate more efficient student learning, each teacher assistant has guided to every other smaller teacher assistant.The existing larger teacher assistants have used to teach a smaller teacher assistant at the next step to increase the learning efficiency.For each mini-batch during training this paper has designed stochastic teaching where a teacher assistant has randomly dropped. The student can always learn rich distilled knowledge from multiple sources ranging from the teacher to multiple teacher assistants.

In paper~\cite{romero2014fitnets} introduced the concept of FitNets for compressing knowledge from thin and deep neural networks to wide networks with more parameters. They have used the advantage of depth in the network compression problem. A part of the teacher's hidden layer is used in the training of the student model, enabling to learn more efficiently. The part from the teacher network is called hints.

\section{Proposed Model}
\subsection{Model architecture}

\begin{figure}[h]
\centering
\includegraphics[width=1\textwidth]{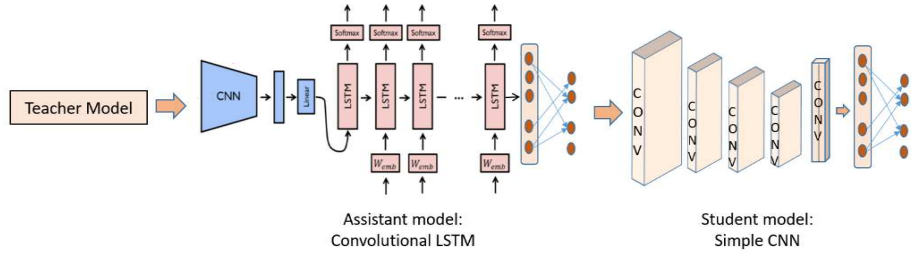}
\caption{Flowchart of our proposed work} 
\label{fig1}
\end{figure}

With increasing complexity of scene text images, state of the art script identification models~\cite{bhunia2019script,shi2016script,gomez2017improving} tend to incorporate more convolutional filters and thereby make a more in-depth and broader network. However, for deploying such models in the practical scenario, it needs to be  pruned or distilled for coping up with available resources. Thus we try to distill the knowledge in various deep neural networks. For the transfer of knowledge, the training is done in three subsequent phases involving an assistant~\cite{son2020densely} and a student network. Figure \ref{fig1} shows the overall model architecture and pipeline to be followed for the knowledge transfer process. This section further describes the details of the three networks mentioned.

\textbf{Teacher Model}
In this literature, we have performed three different experiments involving three various teacher networks. As the teacher network in knowledge distillation theory, is usually taken to be a deep CNN, we have considered three states of the art neural networks pre-trained on the ImageNet dataset~\cite{imagenet_cvpr09} InceptionV3~\cite{szegedy2016rethinking}, VGG19 and EfficientNet-B4~\cite{tan2019efficientnet}. Among these networks, EfficientNet-B4 and InceptionNetv3 have the relatively same amount of parameters which are 17 million and 27 million respectively whereas VGG19 has a vast number of parameters due to more fully connected layers amounting to ~143 million. However VGG has a very simple architecture consisting mainly of linear layers while Efficient Net-B4 and Inception Net V3 have hybrid convolutions in their architectures due to which they have comparatively quite less parameters. Table~\ref{table1} shows the total parameters in the three models discussed. The outputs from the teacher network are divide by a certain temperature value $\tau$, and then softmax is taken to generate soft labels which are used to train the assistant model, as shown in equation \ref{eq:1}.

\begin{table}[!htb]
\caption{Total number of parameters of three different deep neural networks.}
\centering
\resizebox{0.6\textwidth}{!}{
\begin{tabular}{|l|l|l|l|}
\hline
\textbf{Model} & \textbf{Parameters(in millions)}  \\ \hline
InceptionV3   & $\sim$27         \\ \hline
EfficientNet-B4  & $\sim$17         \\ \hline
VGG 19 & $\sim$143         \\ \hline
\end{tabular}
}
\label{table1}
\end{table}

\begin{equation} \label{eq:1}
    \sum_{i}{\frac{\exp(\frac{x_i}{\tau})}{ \sum_{j}{\exp(\frac{x_j}{\tau})}} }
\end{equation}

\textbf{Assistant Model} 
is used as a medium to transfer intermediate features from the teacher to the student network and thus helps in diminishing the feature gap between them. We take the assistant network as a combination of conv-nets and a two-layer  bi-directional stacked LSTM~\cite{bhunia2019script}. Since scene text images contain text instances which are sequential data, recurrent neural networks are the best suited for this purpose. LSTM has gained significant success in the NLU domain and other tasks involving sequential data, and keeping in mind these works, we have incorporated LSTM in our network. There are four convolution layers with 3 x 3 filters and of depth 64, 128, 256, 300 respectively. A max pooling layer follows each of the convolution layers. The result from the conv-net is directed to a two-layer stacked bi-directional LSTM with 256 hidden layers. The results from the LSTM are subjected to two linear layers to give the classified outputs. The assistant network has 14 million parameters which is less than either of the three teacher networks.

\textbf{Student Model} 
is a small network with fewer parameters than the teacher model. The student network learns directly from the trained assistant network. It has undergone training using the soft labels from the teacher network. The student model is a basic conv-net comprising of four convolutional filters and four maxpool layers with 32, 64, 128 and 128 channels respectively. All of them have three kernel size. The final logits are generated through two fully connected linear layers and are used to classify the script in the input images. Owing to the small number of convolutional layers, it has a size of 1 million training parameters, where it is almost 27 times lesser than Inception Network and 143 times lesser than VGG teacher network.


\FloatBarrier

\begin{figure}[h]
\centering
\includegraphics[width=1\textwidth]{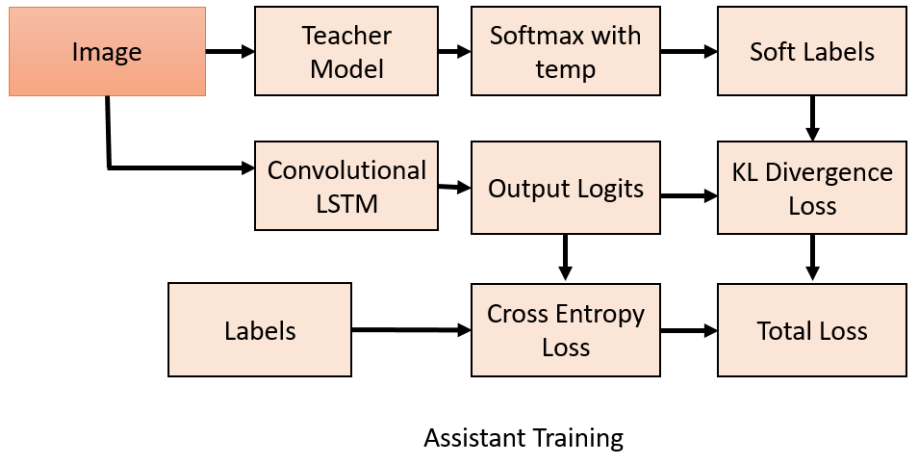}
\caption{Flowchart of assistant model training} 
\label{fig2}
\end{figure}
\FloatBarrier

\begin{figure}[h]
\centering
\includegraphics[width=1\textwidth]{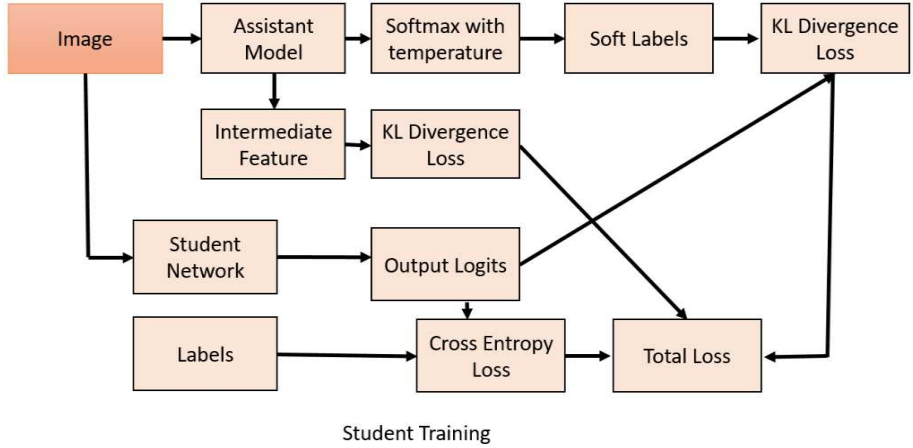}
\caption{Flowchart of student model training} 
\label{fig3}
\end{figure}
\FloatBarrier

\subsection{Training Detail}

\subsubsection{KL Divergence Loss}
compares two probability distributions and generates a similarity score. In this paper, the loss between the predicted results and the soft labels have calculated with the KL Divergence loss.
\begin{equation}
    D_{KL}(p||q) = \sum_{i=1}^{N}{p(x_{i})log(\frac{p(x_{i})}{q(x_{i})})}
\end{equation}

\subsubsection{Distillation Loss}

In knowledge distillation theory, the distillation is taken as sum of mismatch of ground truth label $y^{cls}$  and the outputs from the model penalized by cross-entropy and the mismatch between the soft labels and the student network outputs calculated using KL Divergence loss.
The soft labels are calculated by taking a softmax of the network outputs $\psi$ with a certain temperature $\tau$ .
\begin{equation}
    P^{\tau} = softmax( \psi / \tau )
\end{equation}
For the assistant training step, the distillation loss $K_{DA}$ is given by

\begin{equation}
    K_{DA} = (1 - \alpha)\mathcal{L}_{1}(P_{A}, y^{cls}) + \alpha \tau^{2} \mathcal{L}_{2}(P_{A}^{\tau} ,P_{T}^{\tau})
\end{equation}
where $\mathcal{L}_{1}$ is the cross-entropy loss and $\mathcal{L}_{2}$ is the KL Divergence loss. $P_{A}$ refers to the predicted outputs of the assistant network, and $P_{T}$ refers to that of the teacher network.\\
For the student network, the distillation loss $K_{DS}$ also considers the loss between two similar features vectors between the student and the assistant networks. The first two terms are similar to the cross-entropy loss and the loss between the soft labels and the model outputs in case of the assistant training.

\begin{equation}
    K_{DS} = (1 - \alpha )\mathcal{L}_{1}(P_{S}, y^{cls}) + \alpha \tau^{2} \mathcal{L}_{2}(P_{S}^{\tau} ,P_{A}^{\tau}) + \lambda * \mathcal{L}_{2}(P_{Sf} , P_{Af})
\end{equation}
where $P_{S}$ denotes the predicted results of the student model and $P_{A}$ refers to the assistant network. $P_{Sf}$ refers to the feature vector from the student network and $P_{Af}$ refers to the feature vector from the assistant network. $\mathcal{L}_{1}$ and $\mathcal{L}_{2}$ have the same meanings as in case of the loss of the assistant network.\\
We have taken $\lambda$ and $\alpha$ to be equal to $0.4$ for all experiments.

\subsubsection{Implementation details:}We have used the Pytorch framework for implementation. For training purposes, images were resized to the size of 299 and were transformed to ones with a mean of 0.5 and a standard deviation of 0.5. Augmentations such as random brightness and random contrast were also added to the images during training. We further used the Stochastic Gradient Descent optimizer with nesterov momentum of 0.95 and a learning rate of 0.001 for 150 epochs. We have also used weight decay of 0.001 for training purposes.

\subsubsection{Training}
The training is done on the CVSI~\cite{sharma2015icdar2015} dataset consisting of scene text images. It contains ten classes which are scripts in different languages. The training is a three step process. The first step consists of training of the teacher network. We have studied the performance of three deep neural networks with different architectures - Inceptionet V3, VGG19 and Efficient Net- B4. All three networks have pre-trained on the ImageNet dataset having 1000 classes having a nearly same domain to that of scene text images. The networks are so trained to reduce the loss between the outputs and the ground-truth labels. \\
The second step consists of assistant model to be trained with the distilled knowledge from the trained teacher network. A detailed representation of this step can be found in figure-\ref{fig2}. The assistant network is trained subject to  minimizing the summation of the loss due to predicted outputs with the hard labels and loss between the soft labels generated from the teacher network after taking softmax with temperature and the predicted outputs. To make the training faster, the soft labels were generated before the training and used in the loss function.\\
In the third step, the student network is trained from the knowledge imbibed in the assistant network. The visual representation of this step can be found in figure-\ref{fig3}. The training of the student model is very similar to that of the assistant network with an additional factor. The loss between the feature vector from the assistant network after passing through the lstm module and the feature vector after four convolutional layers in the student network is calculated and added to the distillation loss which is calculated in the same way as in the assistant network.

\section{Experimental Results}
\begin{figure}[h]
\centering
\includegraphics[width=1\textwidth]{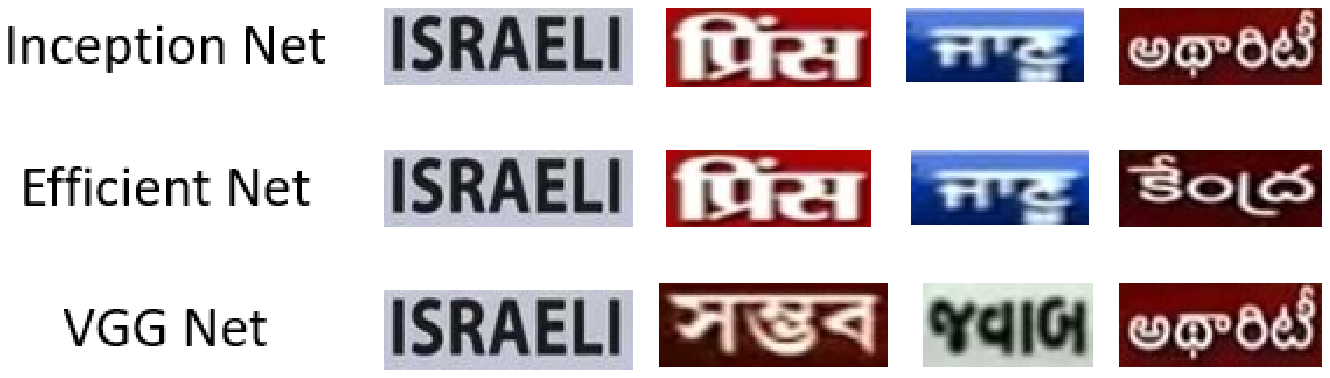}
\caption{Some images of correct prediction} 
\label{fig4}
\end{figure}

\begin{figure}[h]
\centering
\includegraphics[width=1\textwidth]{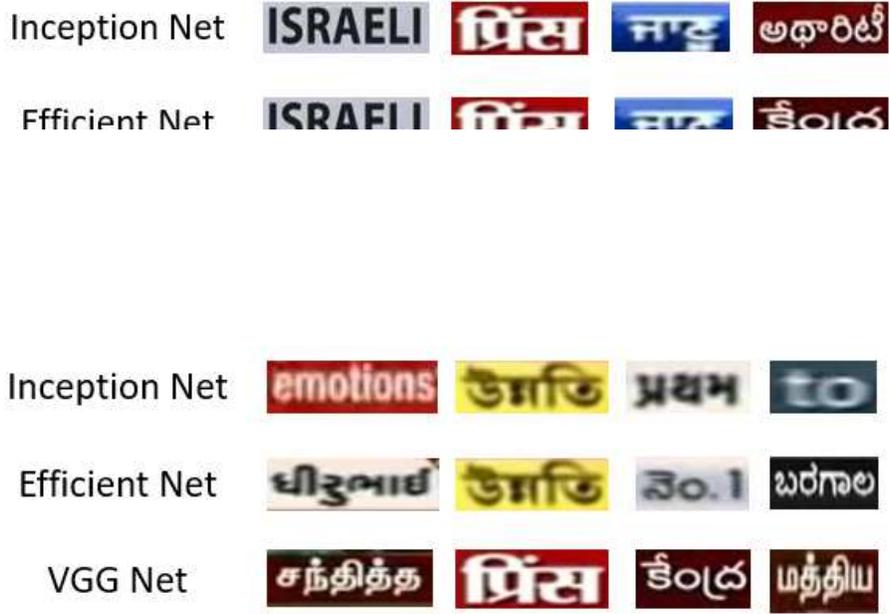}
\caption{Some images of wrong prediction} 
\label{fig5}
\end{figure}

\begin{figure}[h]
\centering
\includegraphics[width=1\textwidth]{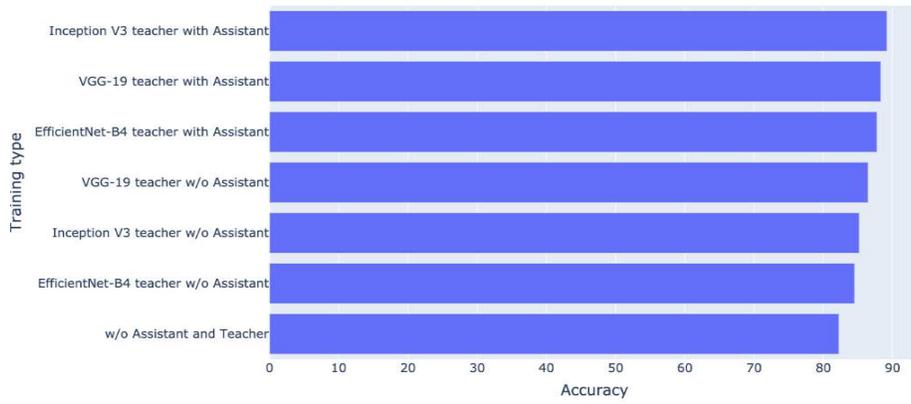}
\caption{A visual representation of the accuracies obtained by the student model trained by different methods of knowledge distillation.} 
\label{fig6}
\end{figure}

We have done several experiments to analyse the performance of our models on the CVSI dataset. CVSI comprises of natural scene text images with different scripts, such as Bengali, English, Hindi, Tamil, Kannada, and other Indian languages. Table ~\ref{table2} shows the final accuracy based on the three teacher networks and the assistant and student networks, trained with knowledge transfer, on the test dataset of CVSI. The teacher network with the highest accuracy enables the highest degree of knowledge transfer in the student network, which is the Inception Net V3 network. For the VGG-19 network, the assistant model has the highest accuracy among the three. Figure~\ref{fig4} shows some correctly identified images due to the three student networks trained from their respective teachers, while figure~\ref{fig5} shows wrongly identified images due to the three student networks. It can be inferred that the Inception Net fails for difficult images due to blurring and colour contrasts whereas in case of VGG it even fails on some simple images. Table~\ref{table3}, Table~\ref{table4}, Table~\ref{table5} shows the precision, recall and f-score related to different scripts in the CVSI dataset of the predictions of the student network trained from the InceptionNet v3, Efficient Net b4 and VGG 19 teacher network. From the tables, it has inferred that Kannada is the most difficult language to identify.
Table~\ref{table6} shows accuracy of the student models based on different experiments performed. The teacher networks were changed and the presence of the assistant network was analysed during the experiments. We can see that the student network learnt better from the VGG as the teacher than inception without an assistant network. This can be due to simple architecture of VGG favouring knowledge distillation than Inception V3. In table~\ref{table6} without the knowledge of the teacher model, the student network alone has the accuracy of 82.24\% on the test dataset. When the student network is trained just on the presence of the teacher network, without the influence of the assistant network, it achieves an accuracy of 85.18 \%. In the presence of  both the teacher and assistant model in subsequent training, the overall model accuracy increases to 89.19 \%. This paper has experimented with three more recent and precise networks like Inception net V3, Efficient Net B4 and VGG-19 as a teacher model. All the performances have done with one strong assistant network Convolutional LSTM. Finally, we have used a small convolutional network as a student model. Figure-\ref{fig6} shows the visual representation of the accuracy obtained by the student network when it is trained with different methods as discussed in table-\ref{table6}. From the results, it can be inferred that with Inception Net V3 as the teacher network, and in the presence of the assistant model, the student model achieves the highest accuracy equal to 89.19\% compared to the accuracies of other student models trained by different methods. 
Table~\ref{table7} shows the script-wise accuracies of different methods due to other researchers and the three different student models trained based on the knowledge distillation from the three different teacher networks - Inception Net V3, VGG19 and Efficient Net-B4. The student model based on the training  from Inception Net V3 has the highest overall accuracy compared to other networks. The highest accuracy corresponding to each script is shown in bold in the table.


\begin{table}[!htb]
\caption{Accuracy chart of teacher,assistant and student model using CVSI-2015 dataset}
\centering
\resizebox{1\textwidth}{!}{
\begin{tabular}{|l|l|l|l|}
\hline
\textbf{Networks}                  & \textbf{Teacher Model} & \textbf{Assistant Model} & \textbf{Student Model} \\ \hline
\textbf{Inception Net V3} & 98.19      & 88.58    & \textbf{89.19}         \\ \hline
\textbf{Efficient Net B4} & 97.19      & 88.52    & 87.74         \\ \hline
\textbf{VGG 19}        & 97.28        & \textbf{89.71}     & 88.30         \\ \hline
\end{tabular}%
}
\label{table2}
\end{table}

\begin{table}[!htb]
\caption{Multi-Script identification results of student model using Inception net v3}
\centering
\resizebox{0.6\textwidth}{!}{
\begin{tabular}{|l|l|l|l|}
\hline
\textbf{Scripts} & \textbf{Precision} & \textbf{Recall} & \textbf{F-Score} \\ \hline
Arabic   & 0.97      & 0.93   & 0.95    \\ \hline
Bengali  & 0.81      & 0.91   & 0.86    \\ \hline
English  & 0.92      & 0.82   & 0.87    \\ \hline
Gujrathi & 0.94      & 0.95   & 0.94    \\ \hline
Hindi    & 0.95      & 0.88   & 0.91    \\ \hline
Kannada  & 0.95      & 0.68   & 0.79    \\ \hline
Oriya    & 0.85      & 0.96   & 0.90    \\ \hline
Punjabi  & 0.95      & 0.93   & 0.94    \\ \hline
Tamil    & 0.84      & 0.95   & 0.89    \\ \hline
Telegu   & 0.79      & 0.90   & 0.84    \\ \hline
\end{tabular}
}
\label{table3}
\end{table}
\FloatBarrier

\begin{table}[!htb]
\caption{Multi-Script identification results of student model using Efficient net b4}
\centering
\resizebox{0.6\textwidth}{!}{
\begin{tabular}{|l|l|l|l|}
\hline
\textbf{Scripts} & \textbf{Precision} & \textbf{Recall} & \textbf{F-Score} \\ \hline
Arabic           & 0.98               & 0.92            & 0.95             \\ \hline
Bengali          & 0.77               & 0.91            & 0.83             \\ \hline
English          & 0.89               & 0.82            & 0.85             \\ \hline
Gujrathi         & 0.97               & 0.93            & 0.95             \\ \hline
Hindi            & 0.95               & 0.86            & 0.90             \\ \hline
Kannada          & 0.97               & 0.58            & 0.72             \\ \hline
Oriya            & 0.89               & 0.95            & 0.92             \\ \hline
Punjabi          & 0.93               & 0.94            & 0.93             \\ \hline
Tamil            & 0.84               & 0.94            & 0.89             \\ \hline
Telegu           & 0.73               & 0.93            & 0.82             \\ \hline
\end{tabular}%
}
\label{table4}
\end{table}
\FloatBarrier

\begin{table}[!htb]
\caption{Multi-Script identification results of student model using VGG 19}
\centering
\resizebox{0.6\textwidth}{!}{
\begin{tabular}{|l|l|l|l|}
\hline
\textbf{Scripts} & \textbf{Precision} & \textbf{Recall} & \textbf{F-Score} \\ \hline
Arabic           & 0.98               & 0.95            & 0.96             \\ \hline
Bengali          & 0.74               & 0.85            & 0.79             \\ \hline
English          & 0.91               & 0.84            & 0.88             \\ \hline
Gujrathi         & 0.93               & 0.96            & 0.94             \\ \hline
Hindi            & 0.87               & 0.88            & 0.87             \\ \hline
Kannada          & 0.97               & 0.65            & 0.78             \\ \hline
Oriya            & 0.80               & 0.99            & 0.88             \\ \hline
Punjabi          & 0.97               & 0.93            & 0.95             \\ \hline
Tamil            & 0.88               & 0.85            & 0.87             \\ \hline
Telegu           & 0.87               & 0.92            & 0.89             \\ \hline
\end{tabular}%
}
\label{table5}
\end{table}
\FloatBarrier

\FloatBarrier

\begin{table}[!htb]
\caption{Comparison of accuracy of the student network trained by different methods}
\centering
\resizebox{0.6\textwidth}{!}{
\begin{tabular}{|l|l|l|l|}
\hline
\textbf{Teacher} & \textbf{Assistant} & \textbf{Accuracy}  \\ \hline
$\sim$          & $\sim$ & 82.24 \\\hline
InceptionV3  & $\sim$ & 85.18 \\\hline
InceptionV3  & Conv-LSTM & 89.19 \\\hline
EfficientNet-B4  & $\sim$ & 84.51 \\\hline
EfficientNet-B4  & Conv-LSTM & 87.74 \\\hline
VGG-19 & $\sim$ & 86.45 \\ \hline
VGG-19 & Conv-LSTM & 88.30 \\ \hline
\end{tabular}%
}
\label{table6}
\end{table}
\FloatBarrier

\begin{table}[!htb]
\caption{A comparative study of the performance of different methods due to different researchers and our student models trained on three different teacher networks based on script-wise accuracies of the CVSI test dataset.}
\centering
\resizebox{1\textwidth}{!}{
\begin{tabular}{|l|l|l|l|l|l|l|l|l|l|l|l|}
\hline
\textbf{Methods}                      & \textbf{English} & \textbf{Hindi} & \textbf{Bengali} & \textbf{Oriya} & \textbf{Gujrati} & \textbf{Punjabi} & \textbf{Kannada} & \textbf{Tamil}  & \textbf{Telugu} & \textbf{Arabic} & \textbf{\begin{tabular}[c]{@{}l@{}}Final \\ Accuracy\end{tabular}} \\ \hline
\textbf{CUK}                          & 65.69            & 61.66          & 68.71            & 79.14          & 73.39            & 92.09            & \textbf{71.66}            & 82.55           & 57.89           & 89.44           & 74.06                     \\ \hline
\textbf{C-DAC}                        & 68.33            & 71.47          & \textbf{91.61}            & 88.04          & 88.99            & 90.51            & 68.47            & 91.90           & 91.33           & \textbf{97.69}           & 84.66                     \\ \hline
\textbf{InceptionV3}          & 82.11   & \textbf{87.73}         & 91.29            & 96.01 & 95.11   & 93.35              & 68.15            & \textbf{94.70}           & 89.16           & 89.78 & \textbf{89.19}                    \\ \hline
\textbf{Efficient-B4}          & 81.52   & 86.19         & 91.29            & 95.09 & 93.27   & \textbf{94.30}              & 57.64            & 93.76           & \textbf{92.57}           & 91.75 & 87.74                    \\ \hline
\textbf{VGG-19}          & \textbf{84.16}    & \textbf{87.73}        & 84.84            & \textbf{99.38} & \textbf{96.02}   &  93.35 & 64.97            & 85.36           & 92.26           & 94.72 & 88.30                    \\ \hline
\end{tabular}
}
\label{table7}
\end{table}
\FloatBarrier

\section{Conclusion}

In this paper, we have studied the effect of knowledge transfer of three different states of the art deep neural networks on a student network for the task of script identification in natural scene text images. We can conclude by our experiments that the teacher network with the highest accuracy, which in our case is Inception Net v3, leads to better knowledge transfer in a student network. Through our experiments, we can also conclude that some languages have difficulties due to colour contrasts and fonts Kannada being one of them. Future work on this will be to improve the assistant and the student networks so that a better knowledge transfer is achieved. 

%
%
%
\bibliographystyle{splncs04}
\bibliography{darkcite}

\end{document}